\documentclass[letterpaper]{article} % DO NOT CHANGE THIS
\usepackage{aaai2026}  % DO NOT CHANGE THIS
\usepackage{times}  % DO NOT CHANGE THIS
\usepackage{helvet}  % DO NOT CHANGE THIS
\usepackage{courier}  % DO NOT CHANGE THIS
\usepackage[hyphens]{url}  % DO NOT CHANGE THIS
\usepackage{graphicx} % DO NOT CHANGE THIS
\usepackage{natbib}  % DO NOT CHANGE THIS AND DO NOT ADD ANY OPTIONS TO IT
\usepackage{caption} % DO NOT CHANGE THIS AND DO NOT ADD ANY OPTIONS TO IT
\usepackage{algorithm}
\usepackage{algorithmic}

\usepackage{newfloat}
\usepackage{listings}

\usepackage{colortbl}
\definecolor{Gray}{gray}{0.92} 
\usepackage{subcaption}
\newcommand{\N}[3]{\mathcal{N}\!\bigl(#1;#2,#3\bigr)}   % compact Normal

\usepackage{mathtools}
\usepackage{enumitem}
\usepackage{makecell}
\usepackage{multirow}
\usepackage{diagbox}
\usepackage{pifont}
\usepackage{booktabs}
\usepackage{tabularx}  
\usepackage{graphicx} 
\usepackage{dblfloatfix}
\newcommand{\cmark}{\ding{51}}  % check mark
\newcommand{\xmark}{\ding{55}}  % x mark
\newcolumntype{Y}{>{\centering\arraybackslash}X}
\newcolumntype{C}[1]{>{\centering\arraybackslash}p{#1}}

\usepackage{amsmath}
\usepackage{amssymb}
\usepackage{tcolorbox}
\usepackage{xcolor}

\usepackage{graphicx}
\usepackage{tabularx}

\definecolor{navy}{RGB}{0, 0, 128}
\definecolor{brickred}{RGB}{203, 65, 84}
\definecolor{darkgreen}{RGB}{0, 100, 0}
\DeclareCaptionStyle{ruled}{labelfont=normalfont,labelsep=colon,strut=off} % DO NOT CHANGE THIS
\floatstyle{ruled}
\newfloat{listing}{tb}{lst}{}
\floatname{listing}{Listing}
\title{Confidence-Guided Stepwise Model Routing for Cost-Efficient Reasoning}
\author{
    Sangmook Lee\equalcontrib,
    Dohyung Kim\equalcontrib,
    Hyukhun Koh,
    Nakyeong Yang, 
    Kyomin Jung\thanks{Corresponding author.}
}
\affiliations{
    Seoul National University\\
    \{helmsman, kimdohyung, hyukhunkoh-ai, yny0506, kjung\}@snu.ac.kr
}

\usepackage{bibentry}
\begin{document}

\maketitle

\begin{figure*}[t]  
  \centering
  \includegraphics[width=\textwidth]{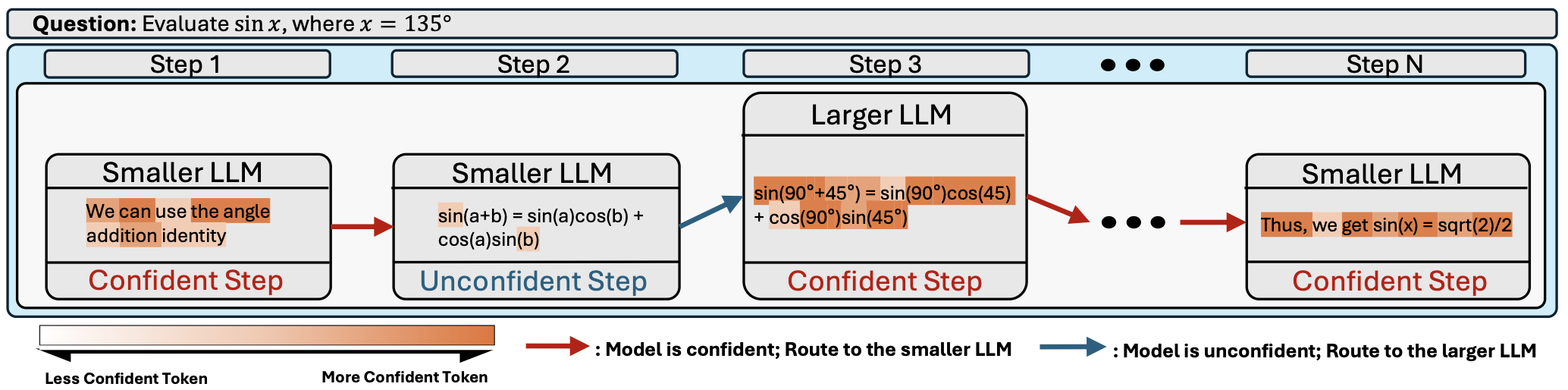}
  \vspace{-0.7cm}
  \caption{\textbf{STEER} uses model confidence to route between the smaller LLM and the larger LLM in a step-wise fashion to achieve a good balance between performance and inference costs. More specifically, the larger LLM is invoked for the generation of the next step only if the smaller model is unconfident in generating the next step. }
  \label{fig:framework}
  \vspace{-0.5cm}
\end{figure*}

\begin{abstract}
Recent advances in Large Language Models (LLMs) - particularly model scaling and test-time techniques - have greatly enhanced the reasoning capabilities of language models at the expense of higher inference costs. To lower inference costs, prior works train router models or deferral mechanisms that allocate easy queries to a small, efficient model, while forwarding harder queries to larger, more expensive models. However, these trained router models often lack robustness under domain shifts and require expensive data synthesis techniques such as Monte Carlo rollouts to obtain sufficient ground-truth routing labels for training. 
In this work, we propose Confidence-Guided \textbf{Ste}pwise Model Routing for Cost-\textbf{E}fficient \textbf{R}easoning (STEER), a domain-agnostic framework that performs fine-grained, step-level routing between smaller and larger LLMs \textit{without} utilizing external models. STEER leverages confidence scores from the smaller model’s logits prior to generating a reasoning step, so that the large model is invoked only when necessary. 
% added for abstract submission
Extensive evaluations using different LLMs on a diverse set of challenging benchmarks across multiple domains such as Mathematical Reasoning, Multi-Hop QA, and Planning tasks indicate that STEER achieves competitive or enhanced accuracy while reducing inference costs (up to $+20\%$ accuracy with $48\%$ less FLOPs compared to solely using the larger model on AIME), outperforming baselines that rely on trained external modules.
Our results establish model-internal confidence as a robust, domain-agnostic signal for model routing, offering a scalable pathway for efficient LLM deployment.

% [Example empirical results in math tasks. Show our work outperforms trained baselines although we are supervision-free. Show that our work maintains the larger model's performance][Example empirical results in ood tasks. Show that our approach is robust] .
% Our findings demonstrate that logits offer a strong and reliable signal for step-level routing in reasoning tasks, paving the way for a more sustainable deployment of LLMs.

\end{abstract}


\section{Introduction}
Recent LLMs demonstrate remarkable abilities in complex, non-trivial tasks such as mathematical reasoning, multi-hop reasoning, and code generation. Their rapid progress can be largely attributed to scaling laws \citep{kaplan2020scaling,DBLP:conf/icml/SardanaPDF24}, in which language models obtain power-law improvements with increased model parameters and data. Exploiting such phenomena has pushed current LLMs into the billion-parameter regime.

 However, such exploitation of scaling laws also led to expensive inference costs, as FLOPs per inference token scale with parameter count \cite{kaplan2020scaling,DBLP:conf/icml/SardanaPDF24}. Furthermore, in the context of LLM reasoning, the problem of expensive deployment is exacerbated by test-time techniques such as Self-Consistency \citep{wangself}, Tree Search \citep{yao2023tree}, and more recently, extended test-time computation using RL techniques \citep{guo2025deepseek} - all of which substantially increase the number of tokens generated per reasoning query. For sustainability and scalability of LLM deployment, addressing the escalating cost of LLM inference has become a timely and pressing challenge \citep{faiz2024llmcarbon,poddar-etal-2025-towards}.

To mitigate such issues, a promising research direction in the domain of cost-efficient inference is adaptive inference using multiple models, where simple questions are allocated to small, cheaper models, and complex questions are allocated to larger, more expensive models. Previous works \citep{chenfrugalgpt,gupta2024language,ong2024routellm,ding2024hybrid,damani2024learninghardthinkinputadaptive} train router models or deferral mechanisms that adaptively allocate an appropriate model at the question-level. While such methods are effective in providing a good quality-cost balance, the routers, being lightweight transformers trained on a specific dataset, show limited performance in out-of-domain inputs \citep{ong2024routellm,ding2024hybrid}. Furthermore, generating the data to train the router typically requires expensive Monte Carlo rollouts to obtain golden router labels  \citep{damani2024learninghardthinkinputadaptive,ding2024hybrid}, or large-scale data-augmentation methods using LLMs \citep{ong2024routellm}. We argue that, although the router may be lightweight, the computational cost of data collection required to train the router is substantial and should not be dismissed as a sunk cost.

In the narrower domain of cost-efficient LLM reasoning, recent works utilize a more fine-grained, step-level signal: RSD \citep{liaoreward} leverages Process Reward Models (PRMs) to guide step-level routing through rejection sampling. Although such works offer fine-grained signals that are well-aligned with the step-by-step reasoning paradigm of LLMs, the reliance on trained PRMs limits applicability predominantly to the domain of mathematical reasoning \cite{zeng2025versaprm} and to the LLMs that the PRM is trained on \citep{zhu-etal-2025-retrieval}. Such drawbacks of previous works that rely on trained external models motivate an external model-free framework that is robust to domain shifts. %Another major drawback of using PRMs for cost-aware adaptive inference lies in the prohibitively high data collection and training overhead. Constructing the synthetic supervision data  for training PRMs involves Monte Carlo sampling to assess step-level correctness for every training query. PRM studies typically generate hundreds of thousands to millions of full reasoning traces, incurring substantial computational cost solely to construct the training dataset \citep{wang2023math,luo2024improve,zhang2025lessons}. Training the PRMs, which utilize LLM backbones, on such large-scale data incurs a commensurately large amount of additional compute. We argue that the computational cost of data collection and model training is too substantial to be dismissed as a sunk cost.%

In this paper, we propose Confidence-Guided \textbf{Ste}pwise Model Routing for Cost-\textbf{E}fficient \textbf{R}easoning (STEER), a novel domain-agnostic, external model-free framework. STEER leverages logit-based confidence scores to dynamically route between smaller and larger models at the step level, reducing inference costs while preserving the reasoning performance of the larger model. This is motivated by the observation that model confidence is strongly correlated with output correctness~\citep{wang2024chain, ma-etal-2025-cot, ma2025estimatingllmuncertaintyevidence}. To assess each model’s reliability throughout the reasoning process, we calibrate these confidence scores using a Gaussian Mixture Model (GMM). As illustrated in Figure~\ref{fig:framework}, STEER evaluates model confidence at every step to determine whether the smaller model is sufficiently capable of handling the current reasoning step, or whether the complexity of the task necessitates invoking the larger model. Notably, in contrast to prior works, STEER provides fine-grained, step-level routing signals \textit{without} relying on external supervision or the training of auxiliary models such as PRMs. This enables broad applicability across diverse reasoning tasks, without the need for additional data collection or fine-tuning. 

Through extensive experiments, we demonstrate that STEER achieves significantly reduced inference cost with minimal performance degradation across diverse domains. Furthermore, STEER is robust to changes in the backbone LLM family, allowing it to generalize effectively across different architectures, substantially broadening its practical utility compared to prior works.
% our approach builds on the Uncertainty Quantification approach by \citet{ma2025estimatingllmuncertaintyevidence}, employing an efficient routing algorithm based on the Gaussian Mixture Model (GMM) formulation to better estimate the utility of calling the larger model. Specifically, [...brief overview]

 %[Empirical results show that ... ]

\subsubsection{Contributions}
Our contributions are as follows:
\begin{itemize}
    \item We propose STEER, a domain-agnostic, external model-free framework for cost-efficient reasoning. Our novel framework leverages logit-based confidence estimation to adaptively route between a larger model and a smaller model at each reasoning step, at test time.
    \item We propose a posterior-based routing strategy leveraging mixture models further to enhance the effectiveness of confidence estimation for routing.
    \item We conduct extensive experiments across a diverse set of benchmarks, including MATH500, AIME, OmniMath, ACPBench, MuSiQue, and KOR-Bench. Across the benchmarks, compared to the larger model, STEER reduces inference costs by 10\% to 48\%, and on AIME, STEER achieves $+20\%$ accuracy with $48\%$ less FLOPs.
\end{itemize}

\begin{figure*}[t]  
  \centering
  \includegraphics[width=\textwidth]{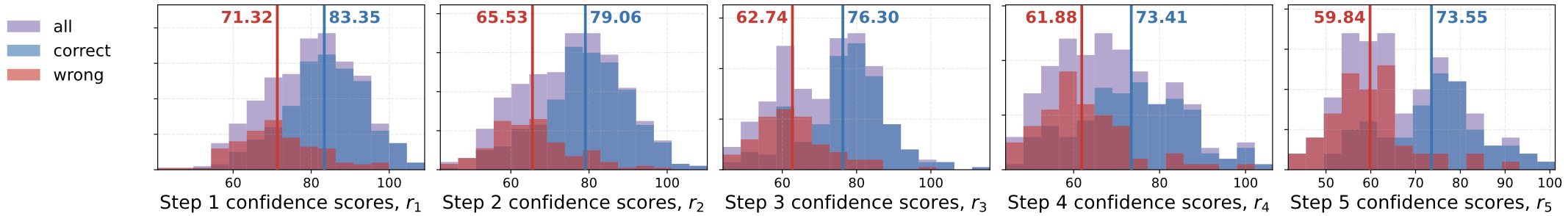}
  \vspace{-0.7cm}
  \caption{Distributions of the confidence scores on correct and wrong reasoning trajectories for MATH500 on Gemma-3-Instruct 4B. The means of correct and wrong cases are shown with respective colors.}
  \label{fig:gmm_math500}
  \vspace{-0.4cm}
\end{figure*}

\section{Preliminaries}
\subsection{Step-Level Model Switching Formulation}
In the context of reasoning, an LLM is fed an input question $x$ and iteratively generates a sequence of reasoning steps $s_1, \ldots, s_n$, which we denote as $s_{1:n}$. Each step $s_i$ is composed of $k_i$ tokens. Formally, the step-by-step generation of the reasoning steps using autoregressive models can be described by the following equation:
\begin{equation}
    s_i \sim p_\theta(\cdot \mid x, s_{1:i-1}),
\end{equation}
\noindent where $p_\theta$ represents the probability distribution from an LLM with parameter $\theta$.

For computational efficiency, works such as RSD \citep{liaoreward} and SpecReason \citep{pan2025specreason} utilize a step-level model switching formulation. These works involve adaptively assigning an larger, more capable model or a smaller, less capable model for the generation of each step. Formally, under this formulation, the generation of the reasoning steps can be described by:
\begin{equation}
    s_i \sim p_{\theta_i}(\cdot \mid x, s_{1:i-1}),
\end{equation}

\noindent where $\theta_i\in \{\theta_M, \theta_m\}$ defines the model chosen for the generation of step $i$, and $\theta_M$ and $\theta_m$ represents the larger model and the smaller model, respectively. In this work, we make use of a GMM that take in as input model confidence scores to choose between $\theta_M$ and $\theta_m$ at every step, invoking the larger model only when necessary in a step-wise fashion.

\subsection{Gaussian Mixture Models}

A two-component Gaussian mixture for a scalar variable \(r\) has density
\begin{equation}
\begin{aligned}
P(r \mid \boldsymbol{\Theta})
  &= \pi_{1}\,\N{r}{\mu_{1}}{\sigma_{1}^{2}}
     + \pi_{2}\,\N{r}{\mu_{2}}{\sigma_{2}^{2}},\\[4pt]
\pi_{1} + \pi_{2} &= 1,
\end{aligned}\label{eq:gmm}
\end{equation}
where \(\boldsymbol{\Theta}=
\{\pi_{1},\pi_{2},\mu_{1},\mu_{2},\sigma_{1}^{2},\sigma_{2}^{2}\}\)
, with \(\pi_{1}, \pi_2 \ge 0\) and \(\sigma_{1}^{2}, \sigma_{2}^{2}>0\). The parameters of the GMM can be obtained using the EM Algorithm. Details on the EM algorithm are in Appendix E.

\paragraph{Posterior assignment}
Let \( z \in \{1,2\} \) denote the latent membership indicator variable. For a new observation \(r\), the posterior probability of membership in
component \(1\) is:
\begin{equation}\label{eq:post_z1}
  P\bigl(z = 1 \mid r,\boldsymbol{\Theta}\bigr)
  \;=\;
  \frac{%
    \pi_{1}\,\N{r}{\mu_{1}}{\sigma_{1}^{2}}%
  }{%
    \pi_{1}\,\N{r}{\mu_{1}}{\sigma_{1}^{2}}
    +\pi_{2}\,\N{r}{\mu_{2}}{\sigma_{2}^{2}}%
  }
\end{equation}
The posterior for component \(2\) is obtained similarly.

\section{STEER}
 We propose a step-level routing framework that adaptively distributes computation between the smaller and larger models. Let $\theta_M$ and $\theta_m$ be parameters of the smaller and the larger models, respectively, and $r(x, s_{1:i})$ denote the reward of a sequence of steps $s_{1:i}$, given a question $x$. Following previous works \citep{liaoreward}, we assume that the larger model outperforms the small model in expected reward: 
\begin{equation}
\mathbb{E}_{s_i \sim p_{\theta_M}} \left[ r(x, [s_{<i}; s_i]) \right] \geq \mathbb{E}_{s_i \sim p_{\theta_m}} \left[ r(x, [s_{<i}; s_i]) \right]
\end{equation}
Under this assumption, routing to the larger model at any given step increases the expected reward. However, under the setting of cost-efficient reasoning, such a routing decision incurs additional computational costs that may be unnecessary. Therefore, an effective routing algorithm must selectively invoke the larger model only when the small model's expected reward is low, aiming to optimize the trade-off between performance and inference cost.

\subsection{Logit-based Confidence Estimation}
Following previous works demonstrating that the correctness of a model is closely related to its confidence \citep{xie2023self, wang2024chain, wangself}, we introduce a confidence-based method to estimate whether a model possesses sufficient capability to generate a correct next step.
Specifically, we leverage logit values as a measure of confidence \citep{wang2024chain, ma2025estimatingllmuncertaintyevidence}. To quantify the confidence of a reasoning step, we first define the token-level confidence score for a single token, and then define the step-level confidence by aggregating token-level scores.
\subsubsection{Token-Level Confidence.}
Let $t_{ij}$ be the $j$-th token of the $i$-th reasoning step. Let $z_{ij} \in \mathbb{R}^V$ denote the vector of vocabulary logits output by a language model when it generates the token $t_{ij}$, where $V$ is the dimension of the vocabulary space of the model. We define the token-level confidence score $\phi_{ij} \in \mathbb{R}^1$ of the token $t_{ij}$ under the language model as follows:
\begin{equation}    
    \phi_{ij} := \max(z_{ij})
\end{equation}
\subsubsection{Step-Level Confidence.}
The stepwise confidence score for step $i$, denoted as $\Phi_i\in \mathbb{R}^1$, is then computed by applying an aggregation function $g$ over the sequence of tokens and their confidence scores associated with step $i$:
\begin{equation}
    \Phi_i := g\left( \phi_{i1}, \phi_{i2}, \dots, \phi_{i{L_i}}\right),
\end{equation}

\noindent where $L_i$ denotes the token length of the $i$-th reasoning step.

The choice of the aggregation function $g$ depends on the reasoning domain. We evaluate STEER on two domains: mathematical reasoning and broader reasoning tasks, such as multi-hop QA and planning. 
In the mathematical domain, we define $g$ as taking the average of the token-level confidence scores over tokens corresponding to mathematical expressions, as mathematical symbols are more critical for solution correctness \citep{lincritical} in mathematical reasoning.
For broader reasoning tasks, we define $g$ as taking the average token-level confidence over all tokens in a step.

% We define the confidence score of a token $t_j$ generated at decoding step $j$, denoted as $c_j$, as the average of its top-$N$ logits:
% \begin{equation}
%     c_t = \frac{1}{N}\sum_{k=1}^{N} z_k,
% \end{equation}

% \noindent where $z_k$ is the $k$-th largest logit associated with token $t_j$.

\subsection{Adaptive Routing via Mixture Models} 
Having defined stepwise confidence scores, we now present our method for adaptive routing between smaller and larger models. Interestingly, distributions of step-level confidence scores for step $i$ for a group of problems, $\mathbf{P}_{\Phi_i}$ (we omit the step index $i$ hereafter in this section for notational simplicity), exhibit a bi-modal distribution, implying that the overall distribution can be decomposed into a mixture two simpler distributions (See Figure \ref{fig:gmm_math500}). Following this observation, we model the distribution of confidence scores as a mixture of two distributions, namely the confident distribution $\mathbf{P}_c$ and the unconfident distribution $\mathbf{P}_u$. 
% In can be inferred from the figure that $\mathbb{E}_{r_i \sim \text{Incorrect}}[r_i] < \mathbb{E}_{r_i \sim \text{Correct}}[r_i]$, underlining the relationship between model confidence and final correctness of a generation. 

More formally, given a group of ongoing responses, at each step, we model the distribution of stepwise confidence scores $\mathbf{P}_\Phi$ as a mixture of two underlying distributions:
\begin{equation}
    \mathbf{P}_\Phi = \pi_c \mathbf{P}_c + \pi_u \mathbf{P}_u
\end{equation}

\noindent where $\pi_c$ and $\pi_u$ are the corresponding mixture weights satisfying $\pi_c+ \pi_u=1$. We also  assume that $\mathbf{P}_c$ and $\mathbf{P}_u$ follow Gaussian distribution, parametrized by $\mu_c, \sigma_c$ and $\mu_u, \sigma_u$, respectively. The formulation separates the reasoning steps into two categories: those that the model can handle confidently, and those that require routing to the larger model to solve it correctly. The parameters $\pi_u, \pi_c, \mu_c, \sigma_c, \mu_u$, and $\sigma_u$ can be estimated by applying the EM algorithm to a group of step-level confidence scores at each step.

Using the fitted $\mathbf{P}_u$ and $\mathbf{P}_c$, we can calculate the posterior probability that an observed step-level confidence score $\Phi$ originates from the confident distribution $\mathbf{P}_c$ over the unconfident distribution $\mathbf{P}_u$. 
In this work, following the observations on the confidence distribution exemplified by Figure \ref{fig:gmm_math500}, we assume %and empirically validate 
that $\mu_u < \mu_c$. We can then obtain a closed-form expression for the posterior probability as follows:
\begin{equation}
    \text{Pr}(c|\Phi) = \frac{\pi_c\mathbf{P}_c(\Phi; \mu_c, \sigma_c)}{\pi_c \mathbf{P}_c(\Phi; \mu_c, \sigma_c) + \pi_u \mathbf{P}_u(\Phi; \mu_u, \sigma_u)}
\end{equation}

\subsection{The STEER Algorithm} 

After estimating the underlying distributions $\mathbf{P}_c$ and $\mathbf{P}_u$, STEER routes the ongoing reasoning traces based on the posterior probability of each confidence score. To be precise, we adopt a posterior threshold $\gamma$ as a hyperparameter for making routing decisions. Given a confidence score of step $i$, $\Phi_i$, the model parameter $\theta_{i+1}$ to be used for generating the next step is determined as follows:

\begin{equation}
    \theta_{i+1} = 
    \begin{cases}
    \theta_m \quad \text{if} \quad \text{Pr}(c|\Phi_i) \geq \gamma \\
    \theta_M \quad \text{otherwise}
    \end{cases}
\label{eqn:threshold}
\end{equation}

\noindent As such, a reasoning trace is routed to a small model if the current step is expected to originate from confident generations. Otherwise, it is routed to the larger model. 

The overall procedure distinguishes between the initial step and all subsequent steps, as no prior confidence score is available for the first step:

\begin{enumerate}[label=\textbf{\arabic*.}, leftmargin=2em, itemsep=0.8ex]
    \item \textbf{Initial Generation}: At the first step ($i=0$), the small model $\theta_m$ generates an output for all questions.
    \vspace{-0.1cm}
    \item \textbf{First-Step Refinement}: Any initial output flagged as unconfident is re-generated with the larger model $\theta_M$.
    \vspace{-0.1cm}
    \item \textbf{Iterative Routing}: For steps $i>1$, each reasoning trace is dynamically routed following the thresholding rule in Equation(\ref{eqn:threshold}). 
    \vspace{-0.1cm}
    \item \textbf{Generation \& Update Confidence}: The selected model generates the current step. Update the step-level confidence scores $\Phi_i$ for the current generation. 
    \vspace{-0.1cm}
    \item \textbf{Iteration}: Repeat steps 3-4 until all questions are solved or reach the maximum iteration limit.
\end{enumerate}
 
\noindent More detailed and formal description of STEER and routing mechanism can be found in Algorithm \ref{alg:STEER} and in Appendix B.

\begin{algorithm}[tb]
\caption{\textsc{STEER}}
\label{alg:STEER}
\textbf{Input}: smaller model $m$ and larger model $M$, their parameters $\theta_m$ and $\theta_M$, questions $\mathcal{X}$, max steps $N$, threshold $\gamma$\\
\textbf{Output}: final solutions $S$
\begin{algorithmic}[1]
    \STATE \text{Generate step and its confidence} $s_{0,m},\Phi_{0, m}\sim p_{\theta_m}(\mathcal{X})$
    \STATE Classify prompts $\mathcal{Q}_M$ to refine with larger model \\ \hfill $(\mathcal{Q}_m,\mathcal{Q}_M)\gets\texttt{classify}(\mathcal{X},{\Phi}_{0,m},\gamma)$
    \FOR{\textbf{each} question $x \in \mathcal{X}$}
        \IF{$x \in \mathcal{Q}_M$} 
            \STATE Refine with larger model $s_{0,M}$, $\Phi_{0,M}$ $\sim p_{\theta_M}(\mathcal{Q}_M)$
            \STATE Add generated steps to prompt $\mathcal{Q}_M \leftarrow \mathcal{Q}_M \oplus s_{0,M}$
        \ELSE
            \STATE Use the step by smaller model $\mathcal{Q}_m \leftarrow \mathcal{Q}_m \oplus s_{0,m}$ 
        \ENDIF
    \ENDFOR
    \FOR{$i = 1$ \textbf{to} $N{-}1$}
        \STATE Route prompts based on confidence \\ \hfill $(\mathcal{Q}_m,\mathcal{Q}_M)\gets\texttt{route}(\mathcal{Q}_m,\mathcal{Q}_M,{\Phi}_{i,m},{\Phi}_{i,M},\gamma)$
        \STATE Generate step and step confidence \\ \hfill
        $s_{i,m}, \Phi_{i,m} \sim p_{\theta_m}(\mathcal{Q}_m)$;
        $s_{i,M}, \Phi_{i,M} \sim p_{\theta_M}(\mathcal{Q}_M)$
        \STATE $\mathcal{Q}_m \leftarrow \mathcal{Q}_m \oplus s_{i,m}$;\quad
               $\mathcal{Q}_M \leftarrow \mathcal{Q}_M \oplus s_{i,M}$
        \IF{$\mathcal{Q}_m$ \textbf{and} $\mathcal{Q}_M$ \text{is all complete}}
            \STATE \textbf{break}
        \ENDIF
    \ENDFOR
    \STATE $S \leftarrow [\,\mathcal{Q}_m;\,\mathcal{Q}_M\,]$
    \RETURN $S$
\end{algorithmic}
\end{algorithm}

\newcolumntype{X}{>{\raggedright\arraybackslash}X}
\newcolumntype{C}{>{\centering\arraybackslash}p{0.2cm}}
\newcolumntype{E}{>{\centering\arraybackslash}m{0.2cm}}
\newcolumntype{W}{>{\raggedright\arraybackslash}p{2.6cm}}  
\newcolumntype{M}{>{\centering\arraybackslash}p{0.75cm}}   
\newcolumntype{G}{>{\centering\arraybackslash\columncolor{Gray}}p{0.80cm}}

\begin{table*}[t]
  \footnotesize
  \centering
  \setlength{\tabcolsep}{2pt}
  \renewcommand{\arraystretch}{1.1}
  \begin{tabularx}{\textwidth}{@{} W E E *{9}{M} | G G G @{}}
    \toprule
    \textbf{Method} &
    \multicolumn{1}{c}{\textbf{Requires Trained}}  &
    \multicolumn{1}{c}{\textbf{Allocation}} &
    \multicolumn{3}{c}{\textbf{MATH500}} &
    \multicolumn{3}{c}{\textbf{OmniMath}} &
    \multicolumn{3}{c}{\textbf{AIME}} &
    \multicolumn{3}{c}{\textbf{Average}} \\[-0.42em]
    \cmidrule(lr){4-6}\cmidrule(lr){7-9}\cmidrule(lr){10-12}\cmidrule(lr){13-15}
    & \multicolumn{1}{c}{\textbf{Ext. Models}} & \multicolumn{1}{c}{\textbf{Granularity}} &
    {\fontsize{7.5pt}{9pt}\selectfont Acc$\uparrow$} & {\fontsize{7.5pt}{9pt}\selectfont FLOPs$\downarrow$} & {\fontsize{7.5pt}{9pt}\selectfont A/F$\uparrow$} &
    {\fontsize{7.5pt}{9pt}\selectfont Acc$\uparrow$} & {\fontsize{7.5pt}{9pt}\selectfont FLOPs$\downarrow$} & {\fontsize{7.5pt}{9pt}\selectfont A/F$\uparrow$} &
    {\fontsize{7.5pt}{9pt}\selectfont Acc$\uparrow$} & {\fontsize{7.5pt}{9pt}\selectfont FLOPs$\downarrow$} & {\fontsize{7.5pt}{9pt}\selectfont A/F$\uparrow$} &
    {\fontsize{7.5pt}{9pt}\selectfont Acc$\uparrow$} & {\fontsize{7.5pt}{9pt}\selectfont FLOPs$\downarrow$} & {\fontsize{7.5pt}{9pt}\selectfont A/F$\uparrow$} \\
    \midrule

    \multicolumn{15}{l}{\textbf{Gemma3-Instruct}}\\[-0.2em]\midrule\midrule
    4B Only         & \hspace*{-2.3cm}-- & \hspace*{-1.4cm}--
            & 73.4 & 8.12 & 9.05  & 24.5 & 12.3 & 1.98 & 10.8 & 11.9 & 0.907 & 36.2 & 10.8 & 3.35 \\
    12B Only        & \hspace*{-2.3cm}-- & \hspace*{-1.4cm}--
            & 84.4 & 19.4 & 4.35  & 33.0 & 35.4 & 0.93 & 17.5 & 33.8 & 0.517 & 44.9 & 29.4 & 1.53 \\
    \midrule
    SpecReason      & \hspace*{-2.3cm}\xmark & \hspace*{-1.4cm}Step
            & 79.8 & 12.3 & 6.45  & 29.6 & 27.0 & 1.10 & 15.0 & 33.6  & 0.447 & 41.4 & \underline{24.2} & \underline{1.71} \\
    \citet{damani2024learninghardthinkinputadaptive}
                    & \hspace*{-2.3cm}\cmark & \hspace*{-1.4cm}Query
            & 85.8 & 17.3 & 4.96  & 32.5 & 29.3 & 1.11 & 15.8 & 34.8  & 0.45 & \underline{44.7} & 27.1 & 1.65 \\
    
    RSD             & \hspace*{-2.3cm}\cmark & \hspace*{-1.4cm}Step
            & 82.8 & 15.1 & 5.50 & 32.9 & 38.6 & 0.85 & 15.8 & 34.4  & 0.460 & 43.8 & 29.2 & 1.50 \\
    STEER  & \hspace*{-2.3cm}\xmark & \hspace*{-1.4cm}Step
            & 85.8 & 15.0 & 5.70  & 33.2 & 26.6 & 1.25 & 15.8 & 30.6  & 0.515  & \textbf{44.9} & \textbf{24.0} & \textbf{1.87} \\
    
    \midrule

    \multicolumn{15}{l}{\textbf{Qwen2.5-Math-Instruct}}\\[-0.2em]\midrule\midrule
    1.5B Only       & \hspace*{-2.3cm}-- & \hspace*{-1.4cm}--
            & 73.0 & 1.82 & 40.1  & 26.4 & 2.72 & 9.7 & 6.67 & 2.94   & 2.28  & 35.3 & 2.50 & 14.1 \\
    7B Only         & \hspace*{-2.3cm}-- & \hspace*{-1.4cm}--
            & 79.6 & 9.10 & 8.75  & 28.2 & 13.56 & 2.08 & 8.33 & 16.6   & 0.500 & 38.7 & 13.10 & 2.96 \\
    \midrule
    SpecReason      & \hspace*{-2.3cm}\xmark & \hspace*{-1.4cm}Step
            & 77.4 & 7.74 & 10.0  & 28.0 & 12.8 & 2.18 & 10.0 & 15.5 & 0.645  & 38.5 & 12.0 & 3.21 \\
    \citet{damani2024learninghardthinkinputadaptive}
                    & \hspace*{-2.3cm}\cmark & \hspace*{-1.4cm}Query
            & 74.8 & 3.08 & 24.3  & 26.8 & 3.33 & 8.05 & 8.33 & 3.10  & 2.69  & 36.3 & \textbf{3.17} & \textbf{11.5} \\
    
    RSD             & \hspace*{-2.3cm}\cmark & \hspace*{-1.4cm}Step
            & 79.8 & 4.56 & 17.5  & 30.0 & 8.54 & 3.51 & 11.7 & 16.9 & 0.692  & \textbf{40.5} & {10.0} & {4.05} \\
    STEER  & \hspace*{-2.3cm}\xmark & \hspace*{-1.4cm}Step
            & 79.6 & 6.38 & 12.4  & 28.3 & 9.16 & 3.09 & 10.0 & 8.64 & 1.16  & \underline{39.3} & \underline{8.06} & \underline{4.88} \\
    \bottomrule
  \end{tabularx}
  \vspace{-0.2cm}
  \caption{Results on mathematial reasoning benchmarks. \textit{Acc} denotes accuracy. \textit{Requires Trained Ext. Models} denote if the method involves employing an external trained model. Best measures are in \textbf{bold}, and the second best \underline{underlined}.}
  \label{tab:main-math}
  \vspace{-0.4cm}
\end{table*}

\newcolumntype{X}{>{\raggedright\arraybackslash}X}
\newcolumntype{C}{>{\centering\arraybackslash}p{0.2cm}}
\newcolumntype{E}{>{\centering\arraybackslash}m{0.2cm}}
\newcolumntype{W}{>{\raggedright\arraybackslash}p{2.6cm}}  
\newcolumntype{M}{>{\centering\arraybackslash}p{0.75cm}}   
\newcolumntype{G}{>{\centering\arraybackslash\columncolor{Gray}}p{0.80cm}}

\begin{table*}[t]
  \footnotesize
  \centering
  \setlength{\tabcolsep}{2pt}
  \renewcommand{\arraystretch}{1.1}
  \begin{tabularx}{\textwidth}{@{} W E E *{9}{M} | G G G @{}}
    \toprule
    \textbf{Method} &
    \multicolumn{1}{c}{\textbf{Requires Trained}}  &
    \multicolumn{1}{c}{\textbf{Allocation}} &
    \multicolumn{3}{c}{\textbf{ACPBench}} &
    \multicolumn{3}{c}{\textbf{MuSiQue}} &
    \multicolumn{3}{c}{\textbf{KOR-Bench}} &
    \multicolumn{3}{c}{\textbf{Average}} \\[-0.42em]
    \cmidrule(lr){4-6}\cmidrule(lr){7-9}\cmidrule(lr){10-12}\cmidrule(lr){13-15}
    & \multicolumn{1}{c}{\textbf{Ext. Models}} & \multicolumn{1}{c}{\textbf{Granularity}} &
    {\fontsize{7.5pt}{9pt}\selectfont Acc$\uparrow$} & {\fontsize{7.5pt}{9pt}\selectfont FLOPs$\downarrow$} & {\fontsize{7.5pt}{9pt}\selectfont A/F$\uparrow$} &
    {\fontsize{7.5pt}{9pt}\selectfont Acc$\uparrow$} & {\fontsize{7.5pt}{9pt}\selectfont FLOPs$\downarrow$} & {\fontsize{7.5pt}{9pt}\selectfont A/F$\uparrow$} &
    {\fontsize{7.5pt}{9pt}\selectfont Acc$\uparrow$} & {\fontsize{7.5pt}{9pt}\selectfont FLOPs$\downarrow$} & {\fontsize{7.5pt}{9pt}\selectfont A/F$\uparrow$} &
    {\fontsize{7.5pt}{9pt}\selectfont Acc$\uparrow$} & {\fontsize{7.5pt}{9pt}\selectfont FLOPs$\downarrow$} & {\fontsize{7.5pt}{9pt}\selectfont A/F$\uparrow$} \\
    \midrule

    \multicolumn{15}{l}{\textbf{Gemma3-Instruct}}\\[-0.2em]\midrule\midrule
    4B Only         & \hspace*{-2.3cm}-- & \hspace*{-1.4cm}--
                   & 53.5 & 5.48 & 9.75 & 43.2 & 0.80 & 54.0 & 46.8 & 2.98 & 15.7 & 47.8 & 3.08 & 15.5 \\
    12B Only        & \hspace*{-2.3cm}-- & \hspace*{-1.4cm}--
                   & 70.7 & 13.5 & 5.20 & 62.6 & 3.80 & 16.4 & 57.0 & 8.96 & 6.35 & 63.4 & 8.78 & 7.20 \\
    \midrule
    SpecReason      & \hspace*{-2.3cm}\xmark & \hspace*{-1.4cm}Step
                   & 62.3 & 12.5 & 4.96 & 55.2 & 2.98 & 18.5 & 52.2 & 8.80 & 5.95 & \underline{56.6} & 8.10 & 7.00 \\
    Damani et al. (2024)
                    & \hspace*{-2.3cm}\cmark & \hspace*{-1.4cm}Query
                   & 67.5 & 11.8 & 5.70 & 47.6 & 1.90 & 25.0 & 52.5 & 6.64 & 7.90 & 55.9 & \underline{6.80} & 8.20 \\
    
    RSD             & \hspace*{-2.3cm}\cmark & \hspace*{-1.4cm}Step
                   & 58.7 & 9.68 & 6.05 & 55.8 & 2.54 & 22.0 & 52.6 & 8.78 & 6.00 & 55.7 & 7.00 & \underline{7.95} \\
    STEER           & \hspace*{-2.3cm}\xmark & \hspace*{-1.4cm}Step
                   & 64.5 & 9.10 & 7.10 & 54.8 & 2.38 & 23.0 & 54.6 & 6.72 & 8.15 & \textbf{58.0} & \textbf{6.06} & \textbf{9.55} \\
    
    \midrule

    \multicolumn{15}{l}{\textbf{Qwen2.5-Instruct}}\\[-0.2em]\midrule\midrule
    1.5B Only       & \hspace*{-2.3cm}-- & \hspace*{-1.4cm}--
                   & 33.1 & 1.20 & 27.6 & 20.0 & 0.76 & 26.3 & 30.5 & 1.06 & 28.7 & 27.9 & 1.00 & 27.9 \\
    7B Only         & \hspace*{-2.3cm}-- & \hspace*{-1.4cm}--
                   & 58.4 & 5.92 & 9.85 & 50.6 & 3.88 & 13.0 & 53.2 & 5.80 & 10.5 & 54.0 & 5.20 & 10.9 \\
    \midrule
    SpecReason      & \hspace*{-2.3cm}\xmark & \hspace*{-1.4cm}Step
                   & 50.3 & 4.36 & 11.5 & 37.8 & 3.22 & 11.7 & 41.4 & 4.62 & 8.95 & 43.2 & \underline{4.06} & \underline{10.6} \\
    Damani et al. (2024)
                    & \hspace*{-2.3cm}\cmark & \hspace*{-1.4cm}Query
                   & 57.8 & 5.98 & 9.65 & 37.4 & 2.52 & 14.8 & 43.7 & 4.64 & 9.40 & \textbf{46.3} & 4.38 & 10.5 \\
    
    RSD             & \hspace*{-2.3cm}\cmark & \hspace*{-1.4cm}Step
                   & 55.3 & 5.08 & 10.9 & 38.4 & 2.90 & 13.2 & 42.6 & 5.20 & 8.20 & 45.4 & 4.40 & 10.3 \\
    STEER           & \hspace*{-2.3cm}\xmark & \hspace*{-1.4cm}Step
                   & 52.6 & 4.16 & 12.6 & 38.6 & 2.72 & 14.2 & 45.2 & 4.02 & 11.2 & \underline{45.5} & \textbf{3.64} & \textbf{12.5} \\
    \bottomrule
  \end{tabularx}
  \vspace{-0.2cm}
  \caption{Results on broader reasoning benchmarks. \textit{Acc} denotes accuracy. \textit{Requires Trained Ext. Models} denote if the method involves employing an external trained model. Best measures are in \textbf{bold}, and the second best \underline{underlined}.}
  \label{tab:broad-reasoning}
  \vspace{-0.4cm}
\end{table*}

\section{Experiments}
\subsection{Experimental Setup}

\subsubsection{Benchmarks.}
To evaluate our method on the mathematical reasoning task, we employ the widely-used (1) MATH500 \citep{hendrycks2021measuringmathematicalproblemsolving,lightman2023let}, (2) Omnimath \citep{gao2024omnimathuniversalolympiadlevel}, and (3) AIME questions collected from 2022 through 2025. MATH500 and OmniMath provide human-annotated difficulty labels, enabling deeper analysis.

Additionally, to evaluate our approach on broader reasoning tasks, we utilize (1) MuSiQue \citep{trivedi2022musique}, a multi-hop QA benchmark, and (2) the MCQ split of ACPBench \citep{kokel2025acpbench}, a benchmark to evaluate LLMs' ability to reason about action and planning. We also use (3) KOR-Bench \citep{ma2025korbench}, and evaluate on  \textit{Cipher}, \textit{Counterfactual}, and \textit{Logic} reasoning subsets\footnote{We exclude \textit{Operation} due to its similarity to math reasoning, and \textit{Puzzles} due to low performance ($<8\%$) across all models.}. More details on benchmarks and evaluation are in Appendix A.

Unless otherwise stated, STEER uses all questions in a benchmark to model $\mathbf{P}_{\Phi_i}$. Specifically, STEER models the distribution from all ongoing reasoning traces across all questions, excluding traces that have already terminated.

\subsubsection{Models.}
We perform experiments on both general-purpose and math-tuned models. Specifically, we evaluate on the general-purpose Gemma3-Instruct \citep{team2025gemma} and Qwen2.5-Instruct \citep{qwen2025qwen25technicalreport} models, and the math-tuned Qwen2.5-Math-Instruct model \citep{yang2024qwen25mathtechnicalreportmathematical}, which was tuned for mathematical reasoning with GRPO \citep{shao2024deepseekmath}. For mathematical reasoning benchmarks, 
we exclude the general-purpose Qwen2.5-Instruct models, as their performance on the AIME benchmark is lacking for meaningful evaluation($<1\%$ accuracy). For broader reasoning benchmarks, we exclude Qwen2.5-Math-Instruct models, as we observed text degradation when applied to non-math tasks. All experiments were performed using vLLM \citep{kwon2023efficient}, and we used \texttt{temperature=0.7} across all settings. We define a reasoning step as a text segment separated by \texttt{\textbackslash n\textbackslash n}.

\subsubsection{Baselines.}
%Better explanations
We adopt two groups of baseline frameworks: \textit{External models}, and \textit{No external models}.

For the \textbf{\textit{External models}} baselines, we adopt (1) RSD \citep{liaoreward} and the question-level model allocation framework by \citet{damani2024learninghardthinkinputadaptive}. In RSD, each reasoning step is first generated using $\theta_m$, after which a PRM assigns a scalar reward to the step. $\theta_M$ is called to regenerate the step only if the reward is below a pre-defined threshold. We use Skywork-o1-PRM 7B\citep{skywork2024}, which is the best-performing PRM in the RSD setting. Note that Skywork-o1-PRM models are trained on Qwen2.5-Math-Instruct models. To the best of our knowledge, there are no PRMs trained on Gemma3-Instruct. (2) \citet{damani2024learninghardthinkinputadaptive} train a lightweight difficulty estimation model to allocate models to input questions efficiently. 
The router takes in the question as input and allocates either $\theta_m$ or $\theta_M$ to generate the entire reasoning trajectory for the query using a thresholding rule. For the \textbf{\textit{No-External Models}} baselines, we adopt the (1) SpecReason \citep{pan2025specreason} design. SpecReason works similarly to RSD, by thresholding step-level reward, but with the reward assignment done by prompting the larger LLM, i.e. by an LLM-Judge. For all baseline methods and STEER, we perform a grid search with a gap of 0.1 over threshold values, selecting the best-performaning value for a fair comparison. 
% Additional details are in Appendix C.
\newcolumntype{Y}{>{\centering\arraybackslash}X}
\subsubsection{Efficiency Metrics}
Following previous works \citep{liaoreward,DBLP:conf/icml/SardanaPDF24,kaplan2020scaling}, as a measure of inference cost, the standard FLOPs estimation for transformers with $N$ parameters - 2$N$ per inference token - is used, and we report the average FLOPs spent per query. We report FLOPs in units of $10^{12}$ for readability. We also report Accuracy-per-FLOPs (denoted as A/F for brevity) adapted from \citet{ma-etal-2025-cot} to better convey the performance–inference cost balance.

\subsection{Results}
The results for mathematical reasoning and broader reasoning benchmarks are presented in Tables~\ref{tab:main-math} and \ref{tab:broad-reasoning}, 
respectively. 

STEER exhibits strong performance across task domains and models. In the mathematical reasoning task, STEER effectively preserves or improves upon the reasoning ability of the larger model, while reducing up to 48\% of the inference cost. Our method remains effective in broader reasoning domains, achieving the best accuracy-cost trade-off across benchmarks. When applied to Gemma3-Instruct models, STEER achieves the highest accuracy with the least FLOPs. With Qwen2.5-Instruct models, our method attains competitive accuracy (within 1\% of the highest-scoring baseline) while using the least (18\% less) computation, further validating the effectiveness of our framework.

\subsubsection{External models have limited robustness}
With Qwen2.5\allowbreak{}-Math\allowbreak{}-Instruct models, RSD attains the highest accuracy in the mathematical reasoning domain. However, with Gemma3-Instruct models, RSD yields lower accuracy on mathematical reasoning benchmarks compared to the larger model, with minimal reduced inference costs. As Skywork-o1-PRM is trained on Qwen2.5-Math-Instruct, the reasoning traces from Gemma3-Instruct present an out-of-distribution problem to the PRM \citep{zhu-etal-2025-retrieval}, thus limiting its ability. Moreover, RSD exhibits sub-optimal performance on broader reasoning tasks, highlighting the limitations of PRMs \citep{zeng2025versaprm} when applied to out-of-domain tasks. We also note that in some cases, RSD incurs more FLOPs than solely querying the larger model. The root cause is the low acceptance rate for the smaller model’s draft steps and the post-hoc evaluation nature of the RSD framework. As the step reward is evaluated post-hoc, rejecting a step leads to invoking \textit{both} the smaller and the larger model. If the acceptance rate is significantly low, both models are invoked for most steps, resulting in FLOPs that match or exceed the sum of FLOPs by the smaller and larger models. 

The query-level router \citet{damani2024learninghardthinkinputadaptive} effectively reduces inference costs on Qwen2.5-Math-Instruct, albeit with visible performance degradation. The performance is sub-optimal with Gemma3-Instruct models across domains, and when run on Qwen2.5-Instruct for broader reasoning tasks, the router routes most queries to the larger model, leading to good accuracy at the cost of heavier inference costs compared to STEER.

\subsubsection{Comparison with LLM-Judge}
STEER outperforms SpecReason across most benchmarks, achieving higher average accuracy while requiring less computation costs. This result demonstrates that prompting is sub-optimal compared to logit-based confidence in providing a performance-inference cost trade-off.

\subsection{Analysis}

\subsubsection{Confidence Measure Selection.}
\begin{figure}[t]  
  \centering
  \includegraphics[width=1.0\linewidth]
  {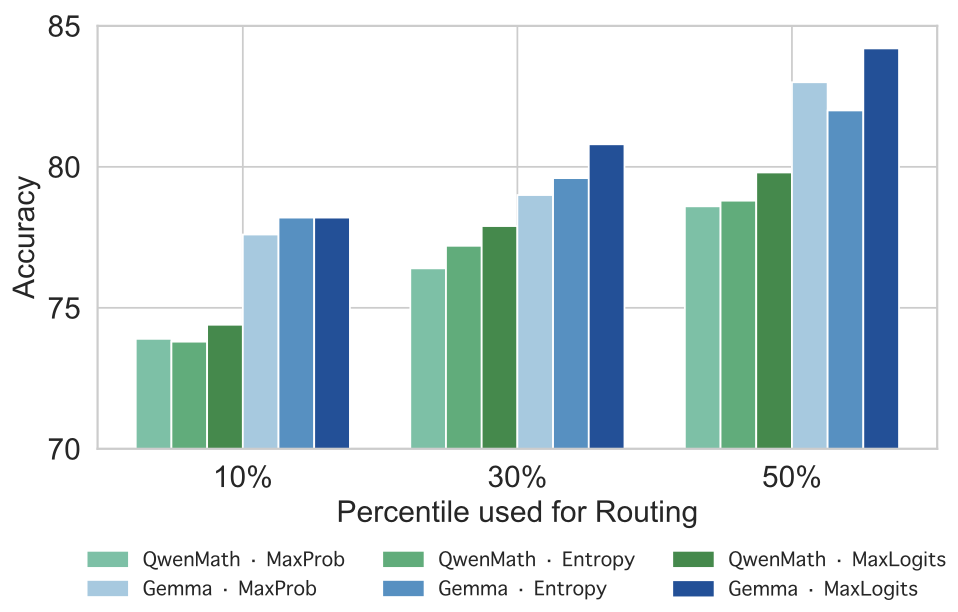}
  \vspace{-0.7cm}
  \caption{MATH500 accuracy for percentile routing with other measures of token confidence. \textit{QwenMath} denotes Qwen2.5-Math-Instruct. \textit{Gemma} denotes Gemma3-Instruct.}
  \label{fig:confi_ablation}
  \vspace{-0.3cm}
\end{figure}

We validate our approach against other widely used confidence metrics, such as entropy and maximum probability. Specifically, we aim to evaluate how effectively each confidence signal identifies incorrect reasoning traces. To this end, we adopt a percentile-based routing strategy. At each reasoning step, we compute the confidence using each metric, and route the traces falling below the $p$-th percentile to the larger model for subsequent step generation. This percentile-based scheme also ensures comparable FLOPs, enabling a fair comparison across metrics. As shown in Figure~\ref{fig:confi_ablation}, maximum logit emerges as the most effective signal, yielding the highest accuracy overall. Notably, when $p$ is small, all confidence metrics perform similarly, suggesting they are equally effective at detecting clearly uncertain cases. However, the superiority of logit-based confidence becomes more apparent at higher values of $p$, indicating its strength in estimating uncertainty even in ambiguous or borderline scenarios.

\begin{figure}
    \centering
    \includegraphics[width=0.8\linewidth]{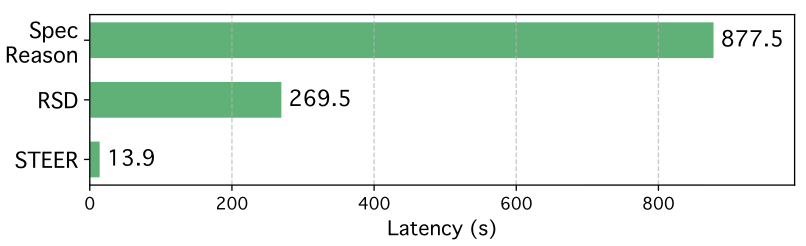}
    \vspace{-0.4cm}
    \caption{Overall latency induced by different methods, measured on the MATH500 benchmark.}
    \label{fig:latency}
    \vspace{-0.8cm}
\end{figure}

\subsubsection{GMM Routing Latency Advantage.}
As shown in Figure~\ref{fig:latency}, the routing latency of GMM employed by STEER is over a magnitude lower than that of the baseline approaches. This can be attributed to the fact that RSD and SpecReason require a full forward pass through an LLM to obtain the routing signals, while STEER utilizes a lightweight GMM.

\begin{figure*}[t]  
  \centering
  \includegraphics[width=\textwidth]{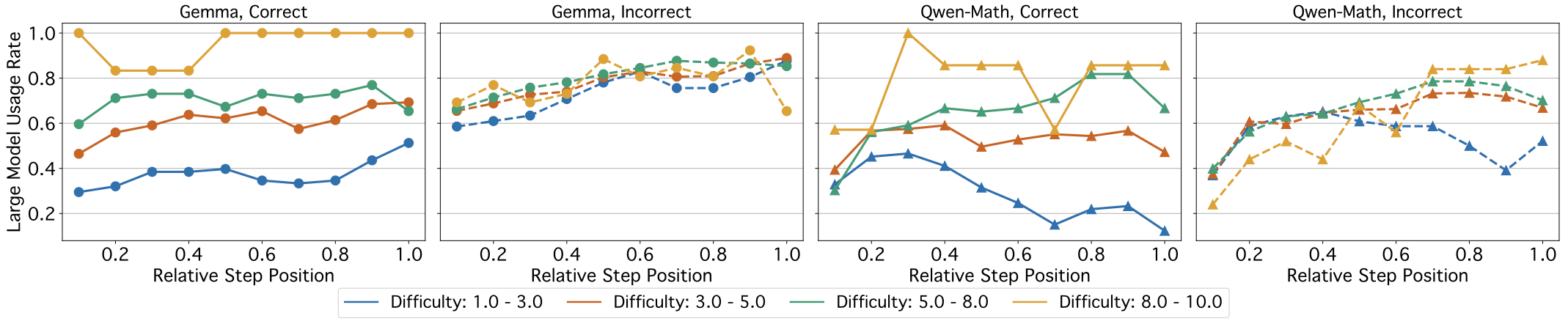}
  \vspace{-0.6cm}
  \caption{Ratio of steps using large models, by relative step position and difficulty in OmniMath using Gemma3-Instruct and Qwen2.5-Math-Instruct models. Relative step 0.1 denotes the first step, and 1.0 denotes the final step. Difficulty of 1.0 denotes the easiest cases, and 10.0 the hardest.}
  \label{fig:usage_pattern}
  \vspace{-0.3cm}
\end{figure*}

\subsubsection{Robustness to Group Size Variations.}
\begin{table}[t]
\footnotesize
\centering
\setlength{\tabcolsep}{6pt}  % 열 간격을 늘림
\renewcommand{\arraystretch}{0.95}
\begin{tabularx}{0.9\columnwidth}{l|l|*{4}{>{\centering\arraybackslash}X}}
\toprule
\multicolumn{2}{c}{\textbf{}} & \multicolumn{4}{c}{\textbf{Number of Groups $K$}} \\
\cmidrule(lr){3-6}
\multicolumn{2}{c}{} & 10 & 5 & 2 & 1 \\
\midrule

\multirow{3}{*}{\textbf{MATH500}} 
& Acc    & 83.9 & 83.5 & 83.5 & 85.8 \\
& FLOPs  & 7.42 & 7.33 & 7.25 & 7.51 \\
& A/F    & 11.3 & 11.4 & 11.5 & 11.4 \\

\midrule

\multirow{3}{*}{\textbf{ACPBench}} 
& Acc    & 63.6 & 64.3 & 64.5 & 64.5 \\
& FLOPs  & 5.04 & 4.80 & 4.53 & 4.55 \\
& A/F    & 12.6 & 13.4 & 14.2 & 14.2 \\

\bottomrule
\end{tabularx}
\vspace{-0.15cm}
\caption{Results on MATH500 and ACPBench using STEER when divided in to K equally sized groups.}
\label{tab:segment_size_comparison}
\vspace{-0.4cm}
\end{table}

We conduct an ablation study where we vary the number of samples used to fit the GMM. In this experiment, a benchmark with $D$ questions is decomposed into $K$ equally sized groups, each with size $D/K$. We solve each group independently using STEER, each using $D/K$ questions to model $\mathbf{P}_{\Phi_i}$. We report the accuracy and FLOPs aggregated across the groups. 
Table~\ref{tab:segment_size_comparison} shows the variation in performance with respect to changes in the group size. Results show that STEER remains effective even when applied to a small-sized group of questions, demonstrating its robustness under limited-sample conditions for confidence distribution estimation.

\subsubsection{Larger Model Usage Pattern.}
We also analyze larger model usage rates by step position and difficulty on the OmniMath benchmark, which has human-annotated difficulty labels ranging from 1 to 10. Figure~\ref{fig:usage_pattern} shows that for both model families, the larger model usage increases with question difficulty in \textit{correct} solutions, implying that STEER correctly leverages more computation when needed. Moreover, large model usage rates are higher in \textit{incorrect} solutions, suggesting that STEER recognizes inability to solve problems correctly and routes more steps to the large model in an effort to solve them. We also note that the usage pattern differs between model families, particularly for correct solutions. Gemma3-Instruct models exhibit relatively consistent usage rates across reasoning steps, while Qwen2.5-Math-Instruct models show divergent usage patterns in later reasoning steps, depending on question difficulty. Still, STEER effectively leverages confidence signals for both model families, validating its robustness across backbone models. 

\section{Related Works}
\subsection{Cost-Efficient Adaptive Inference using Multiple Models}
To reduce inference costs of LLMs, a body of work involves training an external model that allocates models of different sizes at test time. Works on Model Cascading \citep{dohan2022language,chenfrugalgpt,chuang2025learning,narasimhan2025faster} and Model Routing \citep{wang2025mixllm,damani2024learninghardthinkinputadaptive,ong2024routellm} utilize trained models that allocate simple queries to the smaller model and complex queries to the larger model. In the domain of cost-efficient reasoning \citep{sui2025stop}, \citet{liaoreward} propose RSD, which leverages step-level reward values from a trained PRM to perform step-level Model Cascading. However, trained on task-specific and model-specific data, the external models used in these frameworks are not robust to changes in task domains and base models \citep{zeng2025versaprm,zhu-etal-2025-retrieval,ong2024routellm,ding2024hybrid}. 

%Other works focus on reducing latency rather than inference costs.

While STEER focuses on reducing inference costs, other works focus on reducing latency. Speculative decoding \citep{leviathan2023fast,li2025eagle} involves first drafting a set of inference tokens using the smaller model, and then verifying them in parallel by assessing the likelihoods of the draft tokens under the larger model. Although such pipelines can reduce end-to-end latency, they offer no savings in inference compute (FLOPs), as every draft token still undergoes a full forward pass through the larger model during the verification process. 
 
\subsection{Measuring Confidence in Language Models}

Prior works introduce probability-based approaches to measure confidence in generations of a language model, suggesting measures such as maximum probability \citep{arora2021typesoutofdistributiontextsdetect,kim2024personadoubleedgedswordmitigating}, entropy \citep{kadavath2022language}, and semantic entropy \citep{farquhar2024detecting}. 
Recently, \citet{ma2025estimatingllmuncertaintyevidence} 
argues that probability-based measures fail to capture model confidence when equally likely responses exist, and proposes modeling logit-based uncertainty with the Dirichlet distribution for uncertainty quantification.
Other works focus on assessing response-level confidence.  \citet{duan2024shift} argues that not all tokens contribute equally to the semantics of the generated outputs, and suggests taking the tokens with the lowest scores for uncertainty quantification. Similarly, \citet{lincritical} find that certain tokens play more critical roles than others in math reasoning tasks.

\section{Conclusion}
We propose STEER, a novel domain-agnostic, external model-free framework that performs step-level routing for cost-efficient reasoning. STEER leverages logit-based confidence estimation together with GMMs for step-level routing, striking a balance between performance and inference cost. 
Extensive evaluation across reasoning domains and models validate the effectiveness of STEER. By eliminating the need for trained external modules, STEER achieves superior robustness over existing baselines, presenting a practical solution to the challenge of efficient LLM deployment.

\section{Acknowledgments}
This work was supported by Institute of Information \& communications Technology Planning \& Evaluation(IITP) grant funded by the Korea government(MSIT) [No.RS-2023-00229780, Development of Artificial Intelligence Technology for Process-focused Evaluation(Student’s Learning Diagnosis)]. This work was partly supported by Institute of Information \& communications Technology Planning \& Evaluation (IITP) grant funded by the Korea government (MSIT) [No.RS-2022-II220184, Development and Study of AI Technologies to Inexpensively Conform to Evolving Policy on Ethics \& No.RS-2021-II212068, Artificial Intelligence Innovation Hub (Artificial Intelligence Institute, Seoul National University) \& No.RS-2021-II211343, Artificial Intelligence Graduate School Program (Seoul National University)]. In addition, we thank LYWAY for providing math test sets and for the experimental support via API, which contributed to the development of this work. K. Jung is with ASRI, Seoul National University, Korea.

% \section{Proofreading Your PDF}
% Please check all the pages of your PDF file. The most commonly forgotten element is the acknowledgements --- especially the correct grant number. Authors also commonly forget to add the metadata to the source, use the wrong reference style file, or don't follow the capitalization rules or comma placement for their author-title information properly. A final common problem is text (expecially equations) that runs into the margin. You will need to fix these common errors before submitting your file.

\bibliography{aaai2026}

\end{document}